\documentclass{article} 
\usepackage{preprint,times}

\usepackage{amsmath,amsfonts,bm}

\def\eqref#1{equation~\ref{#1}}

\def\1{\bm{1}}

\DeclareMathAlphabet{\mathsfit}{\encodingdefault}{\sfdefault}{m}{sl}
\SetMathAlphabet{\mathsfit}{bold}{\encodingdefault}{\sfdefault}{bx}{n}

\usepackage[utf8]{inputenc} 
\usepackage[T1]{fontenc}    
\usepackage{hyperref}       
\usepackage{url}            
\usepackage{booktabs}       
\usepackage{amsmath}       
\usepackage{amsfonts}       
\usepackage{nicefrac}       
\usepackage{microtype}      
\usepackage{xcolor}         
\usepackage{graphicx}
\usepackage{todonotes}
\usepackage{multirow}
\usepackage{enumitem}
\usepackage[ruled,vlined]{algorithm2e}
\usepackage{dsfont}
\usepackage{tablefootnote}

\newcommand{\name}{ArcMemo\xspace}

\title{ArcMemo: Abstract Reasoning \\ Composition with Lifelong LLM Memory}

\author{%
Matthew Ho$^1$, Chen Si$^1$, Zhaoxiang Feng$^1$, Fangxu Yu$^2$,Yichi Yang$^1$ \\
\textbf{Zhijian Liu$^1$, Zhiting Hu$^1$, Lianhui Qin$^1$} \\
$^1$University of California, San Diego, $^2$University of Maryland
}

\iclrfinalcopy 

\begin{document}

\maketitle

\begin{abstract}

While inference-time scaling enables LLMs to carry out increasingly long and capable reasoning traces, the patterns and insights uncovered during these traces are immediately discarded once the context window is reset for a new query.
External memory is a natural way to persist these discoveries, and recent work has shown clear benefits for reasoning-intensive tasks.  
%
We see an opportunity to make such memories more broadly reusable and scalable by moving beyond instance-based memory entries (e.g. exact query/response pairs, or summaries tightly coupled with the original problem context) toward \textbf{concept-level memory}: reusable, modular abstractions distilled from solution traces and stored in natural language.
%
For future queries, relevant concepts are selectively retrieved and integrated into the prompt, enabling test-time continual learning without weight updates.  
Our design introduces new strategies for abstracting takeaways from rollouts and retrieving entries for new queries, promoting reuse and allowing memory to expand with additional experiences. 
%
We evaluate on ARC-AGI, a benchmark that stresses compositional generalization and abstract reasoning, making it a natural fit for concept memory.
Our method yields a \textbf{7.5\%} relative gain over a strong no-memory baseline, with performance continuing to scale with inference compute.
We find abstract concepts to be the most consistent memory design, outscoring the baseline at all tested inference compute scales. 
%
Moreover, dynamically updating memory during test-time outperforms fixed settings, supporting the hypothesis that accumulating and abstracting patterns enables further solutions in a form of self-improvement.
\footnote{Code available at \url{https://github.com/matt-seb-ho/arc_memo}}.

\end{abstract}

\section{Introduction}
\label{sec:introduction}

Large language models (LLMs) have made substantial strides in reasoning-intensive tasks with long-form reasoning.
However, a notable opportunity lies in the fact that LLM systems are fixed after deployment: new patterns and strategies uncovered during deep reasoning are not yet carried forward once the context is cleared.
This contrasts with human approaches to solving complex, compositional reasoning problems, which involve building on prior insights, abstracting patterns, and composing them in new contexts. 
Augmenting LLMs with memory offers a natural solution to retaining and building on their discoveries.

\begin{figure}
    \centering
    \includegraphics[width=1\linewidth]{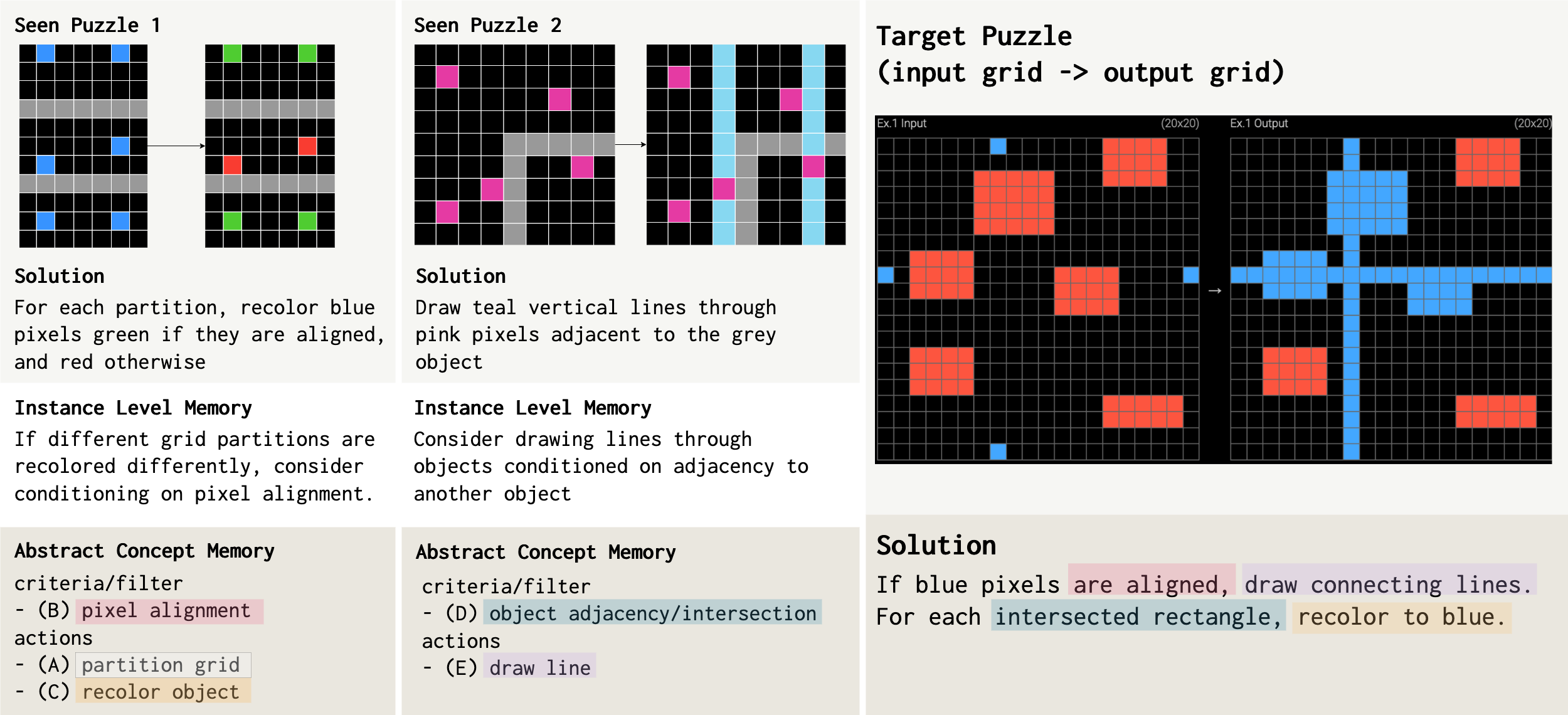}
    \caption{
        \textbf{Instance-Level vs. Abstract Concepts Example.}
        Each ARC-AGI \cite{chollet2019measureintelligence} puzzle requires inferring the transformation rule for a set of input/output pixel grids.
        Here, Puzzle 1 instantiates $(A \land B) \Rightarrow C$ and Puzzle 2 instantiates $D \Rightarrow E$.
        The target puzzle is solved by recombining these ideas ($B \Rightarrow E, D \Rightarrow C$).
        Instance-level memory tends to store fully composed rules, coupling $A$ with $B,C$, and so on. 
        Transferring to the target then demands both ignoring $A$ and disentangling/reordering $B,C$ with $D,E$.
        Abstract memory instead stores $A,B,C,D,E$ as separate, modular concepts, making them easier to recognize and reassemble in new contexts.
    }
    \label{fig:instance_vs_abstract}
    \vspace{-1em} 
\end{figure}

While external augmentation was initially designed for factuality in knowledge-intensive tasks (e.g. \cite{lewis2021retrievalaugmentedgenerationknowledgeintensivenlp} with RAG, \cite{zhang2019ernie} with Knowledge Graphs, \textit{inter alia}), recent work has begun developing memory approaches for reasoning-intensive tasks, storing questions, findings, and mistakes in memory instead of simple facts \citep{yang2024bufferthoughtsthoughtaugmentedreasoning}.
These memories tend to be specific to the problem/experience it was originally derived from (we term these ``instance-level’’ concepts, see \autoref{fig:instance_vs_abstract}). 
Although effective for closely related problems, they have diminished utility for problems superficially different from prior experiences.
Other approaches, such as \cite{suzgun2025dynamiccheatsheettesttimelearning}, begin addressing the ``instance-level’’ issue by maintaining an evolving summary of previous experience to great effect, but the lack of structure and modularity makes scaling with more experiences challenging.
Without selective retrieval, the size of memory is limited, and the problem solving model has the additional burden of picking out the relevant ideas from all past experiences.

We introduce an \textit{abstract concept-level memory} framework to support compositional reasoning for all subsequent queries.
We emphasize 
(1) \textbf{abstract concepts} that are more general and separated from their original context to be useful across a larger set of future problems, and
(2) \textbf{modular concepts} that directly promote recombination with other ideas that can be easily built on with new experiences or curated for a target problem.
We distill solution traces (both from the system itself or from external sources) into reusable, modular abstractions stored in natural language.
At inference time, relevant concepts are selected and integrated into context, enabling test-time continual learning without expensive weight updates.
Our design encapsulates two primary operations:
(i) writing to memory: formatting concepts for abstraction and generality; and 
(ii) reading from memory: selecting a subset of concepts for the current problem.
We title our method \textbf{\name}--- \textbf{A}bstract \textbf{R}easoning \textbf{C}omposition \textbf{Memo}ry.

As illustrated in \autoref{fig:instance_vs_abstract}, instance-level memories often capture an entire solution pattern tied closely to its original problem (e.g., the joint use of ideas $A,B,C$ in Seen Puzzle 1).
Such overspecified entries are less likely to recur in future problems, and even when partially relevant to the Target Puzzle, the agent must still disentangle useful pieces like $B$ from the original bundle.
In contrast, abstract concepts are stored individually with fewer contextual assumptions, making them easier to recognize, adapt, and recombine across superficially different puzzles.

We evaluate on ARC-AGI, where simple pixel-grid operations compose into a vast task space, and solving tasks requires compositional reasoning rather than memorizing individual patterns, making it a natural testbed for {\name}.
On \textsc{ARC-AGI-1}, our method improves the official score from 55.17 to 59.33 (+7.5\% relative gain over a strong no-memory baseline) and is the only memory design we find to outperform the baseline at all inference scales.
Our experiments confirm that continually updating memory at evaluation time yields a score improvement that emerges with inference scale via retries: memory updates triggered from a previous pass over the test set may enable new solves in a subsequent pass. 
Finally, we observe that selecting a subset of memories to include for a particular problem improves performance and reduces token cost, indicating the selection operation is essential beyond allowing memory to grow continually.

\section{Related Work}
Our approach draws on several threads of research in augmenting LLMs with memory but contrasts in (1) target applications (reasoning-intensive vs. knowledge-intensive tasks), (2) the underlying memory modality (text vs. continuous vectors/embeddings), and (3) emphasis on abstraction and modularity for reasoning. 
We discuss related lines (1) and (2) as well as other related work in \autoref{appendix:related_work} and emphasize reasoning and abstraction here.

\paragraph{Memory for Test-Time Reasoning and Concept Abstraction.}
More recent work has shifted toward reasoning-centric uses of memory, particularly test-time learning. 
Think-in-Memory records intermediate reasoning steps as structured triples, storing them in a locality-sensitive hash table to enable reuse in multi-step reasoning tasks \citep{liu2023thinkinmemoryrecallingpostthinkingenable}. 
Buffer of Thoughts stores problem specific reasoning templates, retrieved by embedding similarity and instantiated for new inputs \citep{yang2024bufferthoughtsthoughtaugmentedreasoning}. 
It categorizes and summarizes solution attempts into templated insights, which are then added to memory. 
In contrast, Dynamic Cheatsheet maintains a unified buffer that is adapted continuously. With each problem query, the LLM updates this memory blob by rewriting the entire buffer \citep{suzgun2025dynamiccheatsheettesttimelearning}.
Rather than retrieving specific entries, the entire cheatsheet is appended to the prompt, functioning as a persistent cache of problem-solving strategies.

\paragraph{ARC-AGI Approaches.}
Our work is primarily evaluated against ARC-AGI as a benchmark that simulates complex reasoning on frontier tasks without requiring frontier-level knowledge.
Various techniques and findings have been developed against this benchmark.
\cite{li2024barc} demonstrates the efficacy of synthetic data and the complementary approaches of (i) learning to infer the underlying transformation via program synthesis and (ii) using test-time weight adaptation to directly learn the transformation function in the neural net.
\cite{akyurek2024ttt} further investigates the test-time adaptation method, producing various insights, including that low rank adapters are poorly suited to retaining practical experience across puzzles.
Our program synthesis approach can be viewed as a memory-augmented version of \cite{wang2024hypothesissearchinductivereasoning}'s hypothesis search or \cite{qiu2024phenomenalpuzzlingtestinginductive}'s hypothesis refinement with execution feedback-based retry.
Other recent approaches to ARC-AGI include brain-inspired architectural innovations such as \cite{wang2025hrm}'s hierarchical reasoning model, achieving impressive performance through a recurrent architecture that iteratively refines its output until a halt signal is predicted, together with data augmentation techniques.
Finally, the relative ease of generating new ARC-AGI tasks compared to solving them is exploited in \cite{pourcel2025soararc}'s self-improving system that relabels incorrect programs as the correct program for a different puzzle and trains a model on its generations this way. 

\section{Methods}
\label{sec:method}

\subsection{Problem Statement}

{ \small
\begin{algorithm}[H]
\caption{Inference with Continually Updating External Memory}
\label{alg:inference_memory}
\KwIn{
    Dataset $\mathcal{D} = \{x_i, y_i\}_{i=1}^n$ (labels $y_i$ optional), 
    External memory $M$, 
    Operations \textsc{MemRead}, \textsc{MemWrite}, \textsc{GetFeedback},
    Update interval $k$,
}
\KwOut{Predictions $\{\hat{y}_i\}$}

\For{$i \gets 1$ \textbf{to} $n$}{
    $(x_i, y_i) \gets$ \textsc{GetItem}$(\mathcal{D}, i)$ \tcp*{Label $y_i$ may be absent at inference}
    
    $s_i \gets$ \textsc{MemRead}$(M, x_i)$ \tcp*{Retrieve relevant memory entries}
    
    $\hat{y}_i \gets$ \textsc{LLM\_Generate}$(x_i, s_i)$ \tcp*{Predict using LLM + selected memory}
    
    \If{$i \bmod k = 0$}{
        $f_i \gets$ \textsc{GetFeedback}$(\hat{y}_i, y_i)$ \tcp*{test verification, reflection, etc.}
        
        \textsc{MemWrite}$(M, x_i, \hat{y_i}, f_i)$ \tcp*{Incorporate feedback into memory}
    }
}
\end{algorithm}
}

\begin{figure}
    \centering
    \includegraphics[width=1\linewidth]{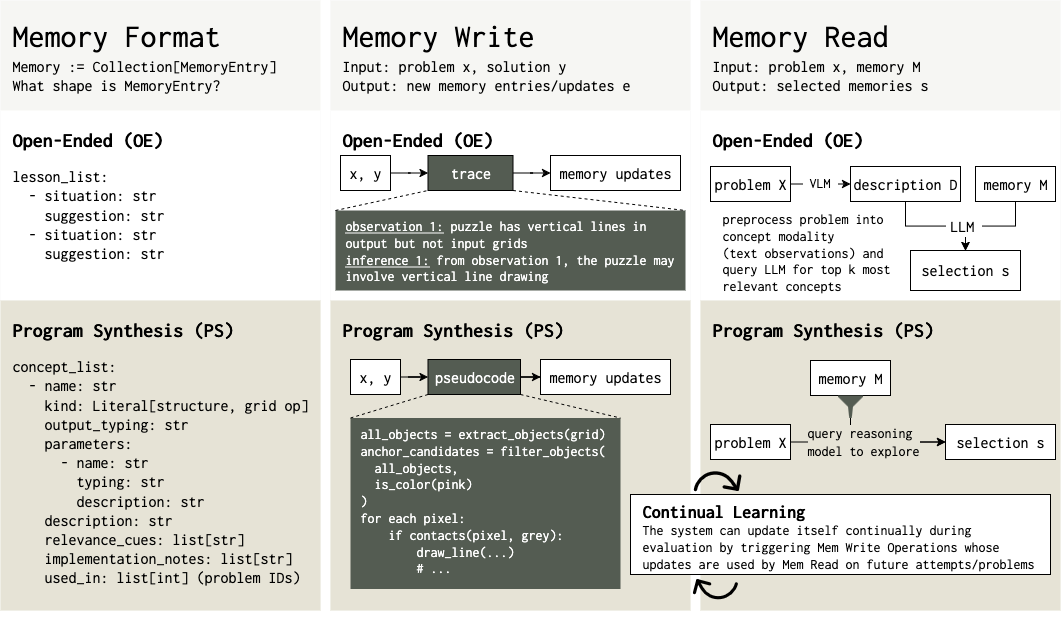}
    \caption{
        \textbf{Method Diagram.}
        Implementing a memory system requires defining (1) what is stored, (2) how memory is updated, and (3) how memory is used for new queries.
        The key novelty in this work is emphasizing abstraction and modularity, and the corresponding design changes.
        In particular, we highlight that parameterization (with higher-order functions allowed and encouraged) promotes abstraction, and typed interface definitions support modularity by showing which concepts can be combined.
        Since these memory entries are more abstract, they also require more inference to map against new, concrete situations--whether by aligning input against the memory format in a preprocessing query, or leveraging reasoning models to explore in a directed manner. 
    }
    \label{fig:method}
    \vspace{-1em} 
\end{figure}

We formulate a memory system as a collection of memory entries associated with read and write operations.
Problem solving with memory augmented LLMs thus requires designing three components as seen in \autoref{alg:inference_memory}:
\begin{enumerate}[itemsep=0mm,topsep=0mm]
    \item \textbf{Memory Format} (\ref{subsec:mem_fmt}): What is stored in individual entries?
    \item \textbf{Memory Write} (\ref{subsec:mem_write}): How is memory updated from a reasoning trace?
    \item \textbf{Memory Read} (\ref{subsec:mem_read}): How is memory used for tackling new problems?
\end{enumerate} 

These are generic design considerations for any memory. 
Our method's novelty lies in the choices made to target (1) abstracting a memory to be less specific to the problem it was derived from (2) enabling memory selection, since greater abstraction makes standard retrieval approaches (embedding similarity thresholding) less effective. 
We present two such implementations: (1) an open-ended (OE) format that imposes minimal constraints on entry format and uses a simple problem preprocessing step for memory selection, and (2) a program-synthesis (PS) inspired format that categorizes and parameterizes concepts and uses reasoning model exploration to select relevant items.

\paragraph{Problem Assumptions.}
The main requirement for our memory system is that some feedback is available at test-time (e.g. via test cases or self-reflection).
This condition is necessary as retaining patterns from an incorrect or otherwise flawed trace would serve to carry mistakes forward.
In other words, some signal is needed to discriminate correct from incorrect traces to ensure only productive ideas and patterns are added to memory.
Incorrect traces still likely contain vital signal, but error identification/credit assignment remains an open challenge that we leave to future efforts.
Various real-world tasks may satisfy this requirement, as tasks are often specified with either examples showing desired behavior or evaluation criteria defining what outcomes are positive and negative.
To give some examples, code completion tools have criteria like compilation success and test pass rate, and medical diagnosis tools have evaluation criteria defined as patient outcomes (e.g., survival rates).
In the case of our evaluation benchmark ARC-AGI, puzzles are explicitly presented as a set of input-output examples, which, for our purposes, can serve as automatically verifiable test cases.

\subsection{Memory Format and Organization}
\label{subsec:mem_fmt}

Specifying an individual memory entry's format involves selecting fields useful for retrieval and downstream problem-solving.
The organization of memory entries on a collective level is another design surface we leave for future exploration; we use a flat collection of entries for simplicity.

\begin{figure}
    \centering
    \includegraphics[width=1\linewidth]{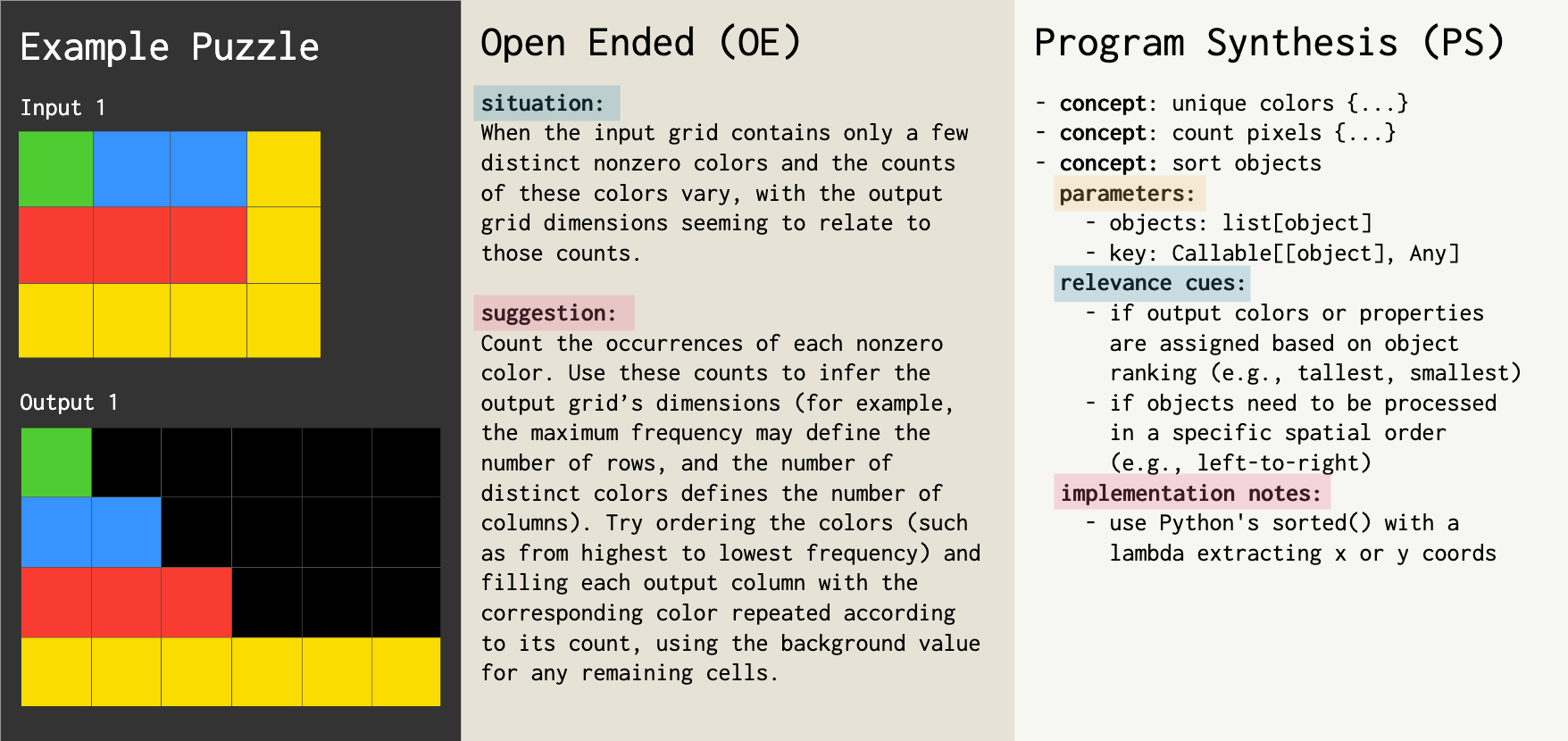}
    \caption{
        \textbf{Open-Ended (OE) vs. Program Synthesis (PS) Concept Examples.}
        An example of concepts from puzzle \texttt{9af7a82c} abstracted into each concept format.
        OE defers to the model, while PS imposes structure to encourage abstraction/modularity.
        Higher order behavior is demonstrated with the ``sort objects'' concept
        taking a \texttt{Callable} parameter that specifies specific variations. 
    }
    \label{fig:oe_vs_ps_ex}
\end{figure}

\paragraph{Open-Ended (OE) Formulation.}
We first consider an open-ended approach that imposes minimal structure and defers formatting individual memories to the model.
Following the basic lesson format of ``under situation X, do action Y'',
the only formatting constraint we impose is having distinct \texttt{situation} (X) and \texttt{suggestion} (Y) fields.
This separation is designed for abstraction (disentangles the core idea from the original context) and modular reuse (\texttt{situation} explicitly describes the conditions where this idea can be reused).

\paragraph{Program Synthesis (PS) Formulation.}
In early iterations of automatic concept summarization from solution traces, we observed outputs to occasionally be overly specific.
Consider \autoref{fig:oe_vs_ps_ex}'s OE example that bundles ideas of counting, drawing columns, and sorting into a single entry.
This overspecification inherits the same limitations as instance-level concepts: friction when matching a composite idea to new scenarios and redundancy from composite ideas sharing components.

We designed a more structured alternative to the OE format to address these issues and more strongly enforce the ideals of abstraction and modularity.
In particular, we took inspiration from software engineering and functional programming, which respectively have existing solutions in modular design for code reuse and higher-level functions for composition.
From this analogy, our Program Synthesis (PS) format frames concepts as types, structures, and routines.
The primary feature is that each of these concepts is parameterized.
Parameterization allows similar concepts to be represented compactly with variations abstracted into parameters.
Moreover, type annotations for inputs (parameters) and outputs (return types) encourage modularity and composition: the typed interfaces suggest what ideas fit together and how.
Finally, by allowing parameters to be routines themselves (introducing higher-order functions), we encourage further generalization by enabling abstraction over specialized logic and routines rather than just values.
Higher-level patterns can be recognized across instances and stored with routine arguments defining the different lower-level logic.
A subtle benefit of this more structured formulation is that memory representation can be easily compressed by omitting certain fields (see \autoref{appendix:ps_mem_fields} for a complete list of PS concept fields).

\subsection{Memory Write: Concept Abstraction}
\label{subsec:mem_write}
Memory's primary purpose is to persist discoveries and reflections from prior experiences.
Converting experiences to memory updates is then the most crucial operation for achieving persistence.

\paragraph{OE Abstraction.}
Abstracting suggestions from a solution trace is straightforward--simply query a model to reflect on the solution trace and summarize specific general ideas that may be reused for future puzzles.
To synthesize specific conditions or situational cues to pair with suggestions, ideally, we can directly refer to the original problem-solving process as a series of deductions that derive new facts or explore ideas from existing facts, and link each suggestion (idea) with the prior observations that inferred this suggestion.
However, intermediate inferences may be implicit or difficult to extract from a trace, or, in the case of specific commercial models, completely hidden.
To address this challenge, we generate a post-hoc derivation: an interleaved sequence of observations and thoughts/reasoning steps constructed from the input-output of the final solution.  
These reconstructed traces are then used to extract situation-suggestion pairs, forming structured memory entries for guiding future analyses.
This minimal formulation does not consider the existing contents of memory while adding new entries for simplicity and scaling reasons.
It defers handling redundancy and consolidating ideas to the concept selection and problem-solving phases, respectively.

\paragraph{PS Abstraction.}
We find that directly converting solutions into routines often leads to minor implementation details being recorded as memories, as opposed to our intended abstract concepts.
We mitigate this by preprocessing solutions into pseudocode to prioritize higher-level operations over low-level implementation details.
The main abstraction phase proceeds from the generated pseudocode, recording new concepts and revising existing concepts along with their various fields.

In the spirit of designing for abstraction, promoting reusability, and minimizing total concept memory length, this operation \textit{is} aware of existing concepts with a compressed form of memory included in context.
We encourage the model to reuse and revise existing concepts by updating concept descriptions, parameter lists, relevance cues, and implementation notes.
Importantly, we encourage higher-order functions with routines as arguments in instructions and few-shot demonstrations.

All concept abstraction/memory writing operations are scaffolded with few-shot demonstrations, example-rich templates, and comprehensive instructions.

\subsection{Memory Read: Concept Selection}
\label{subsec:mem_read}
Although state-of-the-art models support longer context windows, their maximum length still limits how much information or how many memory entries can be included in a single query.
Even without this hard limit, including all available memories in context is still undesirable: it can distract the model with irrelevant details and flood it with too many unlikely hypotheses to consider.
A more effective strategy is to introduce a selection mechanism only to include the most relevant subset of memory entries at problem-solving time.
This allows the memory store to grow continually without overwhelming the model's context window.

\paragraph{OE Selection.}
OE-format memory entries were explicitly designed to support memory-based selection.  
The \texttt{situation} clause acts as a semantic hook to identify relevance.
The key idea for OE Selection is a preprocessing step that leverages model reasoning capability to parse the abstract domain input into problem descriptions at different levels of abstraction to facilitate matching against our abstract concepts.
Since ARC-AGI deals with spatial reasoning, we leverage a vision language model for this preprocessing step. 
We caption each puzzle using a structured prompt that separates concrete observations from speculative transformations.  
This converts spatially rich input into a natural language format suitable for matching against stored situations.
We then query a model for the top-$k$ most relevant entries using the generated description.  
We explored scoring, thresholding, and cumulative similarity approaches (as in top-$p$ sampling).
Observing comparable results, we selected top-$k$ cutoff for simplicity.
We consider this a light-weight approach to converting abstract domain inputs into more understandable and memory-aligned representations.

\paragraph{PS Selection.}
Standard embedding approaches use a single forward pass per embedding, representing a System 1-style fast thinking/intuition-driven understanding from the model.
Intuitions may not be sufficient for difficult/frontier tasks and domains.
Moreover, the connection between a higher-level concept and a specific problem instance may not be immediately clear because of the abstraction.

To address these challenges, we propose \textbf{reasoning-based selection}--using System 2-style thinking to deliberately think and explore.
In contrast to the more straightforward playbooks curated in the OE memory format, the explicitly abstract PS concepts are a set of puzzle pieces with notes on identifying if a piece is relevant (relevance cues) and on how various pieces fit together (type annotations).
To select from this memory, we propose leveraging model-driven exploration instantiated by recent models' long-form reasoning with backtracking.
PS Selection instructs a reasoning model to systematically explore the problem: first identify initial concepts using relevance cue annotations, then attempt to ``fill in the details'' by determining values or routines to populate these initial concepts' parameters using type annotations to identify which other concepts should be investigated.



\subsection{Continual Concept Learning}
A long-held goal in machine learning is to develop lifelong learning systems that continually self-improve without manual intervention.
A memory system that can leverage the learning signal present at test-time via continual updates is one approach to achieve this ambition.
Our memory write operations are lightweight queries that can ingest solution traces derived from both the system and external sources, making continual updates practical at scale.

However, continual updates also introduce dependencies on evaluation order.
If solving problem $i$ induces a memory update that enables problem $j$ to be solved, then performance differs between $(...,x_i,...,x_j,...)$ and $(...,x_j...,x_i,...)$.
Inference batching further complicates this: even if $x_i$ precedes $x_j$, they may appear in the same batch, so the model attempts $x_j$ before $x_i$ has updated memory.
This introduces an accuracy--throughput trade-off.
In all settings, we initialize our external memory with seed data (problems and solutions).
We explicitly evaluate continual updates in \autoref{subsec:continual}'s experiments, confirming their efficacy.
In other experiments, we used fixed memory to prioritize throughput and avoid the potentially confounding effect of order dependencies.

\section{Experiments}
\label{sec:experiments}

\paragraph{Benchmark Selection.}
We evaluate our proposed framework on ARC-AGI-1~\citep{chollet2019measureintelligence}, a benchmark explicitly designed to evaluate intelligence as ``efficient acquisition of new skills'' instead of ``fixed possession/memorization of skills.''
Each ARC puzzle encodes a transformation rule that maps input to output pixel grids.
The objective of each puzzle is to infer its rule given several examples of input-output grid pairs, and produce the corresponding output grid for several input test cases.
We find that the abstract domain of pixel grid transforms provides a meaningfully challenging testbed to simulate frontier domains without requiring expert knowledge to evaluate trajectories--a confluence of desirable properties for evaluating a continual concept learning system.

ARC-AGI-1 contains a public validation split containing 400 puzzles with a difficulty distribution matching that of the private evaluation.
Following \cite{akyurek2024ttt}, we evaluate a randomly selected 100-puzzle subset of the public val split.
This makes repeated runs for more stable estimates feasible given cost and the sampling variance we observed.
\cite{li2024barc} manually authored Python solutions for 160 puzzles from the public train split to act as ``seeds'' to recombine into their synthetic dataset.
We reuse these solutions to seed our memory rather than training data.

\paragraph{Models.}
To build on frontier models, we experiment primarily with OpenAI's o4-mini.
At the time of writing, o4-mini is second only to Grok 4, but o4-mini's lower price puts it on the Pareto frontier of cost and performance \cite{arcprizeleaderboard}.
For auxiliary tasks such as concept abstraction and non-reasoning selection, we use OpenAI's GPT-4.1 to conserve token usage. 
Early experiments also evaluated the open-weight DeepSeek R1, which has the benefit of visible thinking traces, but its $8000$ output token limit consistently yielded unfinished solutions in initial testing.

\paragraph{Evaluation.} 
While the official evaluation harness queries models to directly predict output grids for test cases, we instead use a program synthesis approach that queries for a transformation function to convert input to output grids.
The code artifact provides more signal for reflection and also allows us to test proposed logic against reference pairs for feedback.
We evaluate performance under 0, 1, and 2 retries with this execution feedback.
We follow official ARC-AGI scoring (two attempts per puzzle), and account for sampling variance by averaging over extra runs (see details in \autoref{appendix:eval_details}).
The primary memory baseline we compare against is a re-implementation of \cite{suzgun2025dynamiccheatsheettesttimelearning}'s DC-Cu (labeled ``cheatsheet'') that uses a frozen memory to match other settings.


\section{Results}
\label{sec:results}

\subsection{Main Results}
\label{subsec:main_res}

\begin{table}[ht]
\centering
\small
\begin{tabular}{lrl}
\toprule
Setting & Oracle@1 & \textbf{Oracle@2 (Official)} \\
\midrule
qwen3-235b-a22b-instruct & 11.00 (1.00) & 16.67 (0.57) \\
deepseek r1 & 19.50 (7.05) & 26.33 (4.02) \\
claude sonnet 4 (thinking 16k) & 45.50 (1.00) & 54.17 (0.58) \\
gemini 2.5 flash (thinking 16k) & 9.67 (4.86) & 13.83 (4.73) \\
o4-mini (medium) & 46.33 (1.04) & 55.17 (3.18) \\
cheatsheet (o4-mini (medium)) & 47.50 (2.78) & 57.67 (2.52) \\
\midrule
\name-OE (ours, o4-mini (medium)) & 48.00 (1.00) & 56.67 (1.53) \\
\textbf{\name-PS (ours, o4-mini (medium))} & \textbf{49.33 (0.29)} & \textbf{59.33 (0.29)} \\
\quad + one retry & 58.00 (2.29) & 67.33 (1.61) \\
\quad + two retries & 61.67 (3.88) & 70.83 (3.06) \\
\bottomrule
\end{tabular}

\vspace{0.5em}
\caption{
    \textbf{Main Results.}
    \name-PS yields the best results and scales with additional compute.
    Bracketed values represent standard deviation.
    See additional score details in \autoref{tab:full_res}.
}
\label{tab:main_exp}
\end{table}

As seen in \autoref{tab:main_exp}, \name-PS achieves the best official score with standard compute \footnote{\label{note:standard_compute}Matching the official evaluation scheme with 2 parallel attempts and 0 retries.} and still benefits from additional scaling in both parallel samples and sequential retries.
\autoref{tab:full_res} shows complete details on all settings at all scales, where \name-PS is the only setting that consistently outperforms the baseline in all inference compute scales.
Other memory formulations (Cheatsheet, \name-OE) situationally improve over the baseline but still underperform it in some regimes.

We observe that \name-PS's advantage is most prominent with lower compute regimes.
This is consistent with the stated goal ``memory for reasoning'': persisting previous experience aims to reduce redundancy from rediscovering ideas found in previous rollouts. 
The marginal impact of memory is reduced with more inference compute as the model can rediscover ideas through exploration.

Compared to our memory baseline (Cheatsheet), \name methods are highly competitive, performing favorably in most regimes.
The quantitative improvement shown by \name is supported by qualitative analysis that demonstrates \name memories are more modular.
This qualitative analysis (described more in \autoref{subsec:ablate_select}) observes that these abstract memories improve concept coverage--more target puzzle ideas are reflected in memory.
This difference may be explained by \name's modular design being more conducive to maintaining a growing library.

\subsection{Selection Ablation and Qualitative Relevance Analysis}
\label{subsec:ablate_select}

\begin{table}[ht]
\centering
\small
\begin{tabular}{lllcr}
\toprule
\multicolumn{2}{c}{} &  \multicolumn{3}{c}{Oracle@k Scores} \\
\midrule
 &  & k=1 & \textbf{k=2 (Official)} & k=3 \\
retry & setting &  &  &  \\
\midrule
\multirow[t]{2}{*}{0} & \name-PS & \textbf{49.33 (0.29)} & \textbf{59.33 (0.29)} & \textbf{63.50} \\
 & \quad - selection & 46.83 (5.25) & 55.17 (2.02) & 60.00 \\
\midrule
\multirow[t]{2}{*}{1} & \name-PS & \textbf{58.00 (2.29)} & \textbf{67.33 (1.61)} & 72.50 \\
 & \quad - selection & 57.67 (3.79) & 66.00 (3.00) & \textbf{73.00} \\
\midrule
\multirow[t]{2}{*}{2} & \name-PS & \textbf{61.67 (3.88)} & \textbf{70.83 (3.06)} & 75.50 \\
 & \quad - selection & 61.33 (3.21) & 70.00 (2.65) & \textbf{76.00} \\
\bottomrule
\end{tabular}

\vspace{0.5em}
\caption{\textbf{Selection Ablation.} Ablating the reasoning-based selection mechanism from \name-PS reduces overall performance, showing selection is useful for downstream performance.} 
\label{tab:selection_ablation}
\end{table}

We perform an ablation experiment removing the selection mechanism from \name-PS to examine its impact and to compare more directly to the selection-free Cheatsheet methodology.
The generally better performance of the with-selection setting (as seen in \autoref{tab:selection_ablation}) suggests that, in addition to allowing memory to grow without overwhelming context in both the abstraction and inference phases, it also helps downstream problem-solving performance.
Moreover, from the token usage plot in \autoref{fig:token_efficiency}, we see that concept selection's performance improvement also comes at higher efficiency--the selection-ablated setting uses far more tokens.
However, there still exist certain scale regimes where the no-selection setting has a better score.
We attribute this to variance from imperfect selection and note that this indicates headroom for selection/retrieval.

\name-PS (-selection) is most comparable to Dynamic Cheatsheet's ``cumulative'' setting (DC-Cu) as all stored memory is included with each query.
While both ultimately solve the same number of puzzles across all runs, we find that the settings differ in 10 puzzles (that is, each system solves 5 puzzles the other does not).
We manually analyzed these puzzles and found that only 40\% of new solves in the cheatsheet setting were related to memory elements actually in the generated cheatsheet.
That is, we were unable to find relevant notes in the cheatsheet to explain 60\% of its new solves.
In contrast, 100\% of new solves from \name-PS (-selection) can be linked to concept memory contents.
While we cannot definitively conclude \name changes directly induced each of the new solves (given the state of LLM interpretability), this seems to suggest new solves in the \name settings are potentially more attributable to the memory component compared to sampling variance.

Manual inspection of a subset of reasoning-based selection results used for \name-PS finds that while some irrelevant concepts still appear within the selection, the key idea of the target is often included in the selection.
The presence of a few distractor concepts appears to be manageable, as the reasoning model itself is capable of exploring with backtracking.

\subsection{Continual Learning}
\label{subsec:continual}

\begin{table}[ht]
\centering
\small
\begin{tabular}{lllcr}
\toprule
\multicolumn{2}{c}{} &  \multicolumn{3}{c}{Oracle@k Scores} \\
\midrule
 &  & k=1 & \textbf{k=2 (Official)} & k=3 \\
retry & setting &  &  &  \\
\midrule
\multirow[t]{2}{*}{0} & \name-OE & 48.00 (1.00) & 56.67 (1.53) & 61.00 \\
 & \quad + continual memory update & 46.33 (1.53) & 56.00 (0.00) & 61.00 \\
\midrule
\multirow[t]{2}{*}{1} & \name-OE & 56.67 (2.08) & 65.67 (1.53) & 70.00 \\
 & \quad + continual memory update & 57.17 (3.69) & 65.00 (0.00) & 67.00 \\
\midrule
\multirow[t]{2}{*}{2} & \name-OE & 60.67 (1.53) & 67.67 (2.52) & 71.00 \\
 & \quad + continual memory update & \textbf{62.33 (3.51)} & \textbf{70.00 (1.73)} & \textbf{72.00} \\
\bottomrule
\end{tabular}

\vspace{0.5em}
\caption{\textbf{Continual Memory Learning.} Comparing otherwise identical memory systems, we find that with sequential inference compute scale (at high retry depth), continually updating memory with new self-generated solutions leads to improved puzzle-solving performance.}
\label{tab:continual_mem}
\end{table}

Results from our experiment comparing \name-OE with a version that updates \textit{during} evaluation (every 10 problems; \autoref{tab:continual_mem}) show that our memory system's performance improves with continual updates--a key requirement for meaningful lifelong learning.
In particular, we find that the performance improvement emerges at later iterations in sequential inference scaling.
We hypothesize this is because, after more iterations and puzzle-solving passes, new solutions are found, new memories are abstracted, and these new memories help solve other puzzles. 

We further discuss token efficiency, concept specificity, and embedding retrieval in \autoref{appendix:further_discussion}, and limitations and future work discussion in \autoref{appendix:limitations}.

\section{Conclusion}
\label{sec:conclusion}

In this work, we introduce {\name}, a framework designed to support lightweight, lifelong learning on compositional reasoning tasks by emphasizing a higher level of abstraction and modularity.
We explore two implementations of memory modules against ARC-AGI, a benchmark specifically designed to resist memorization and to evaluate fluid intelligence instead.
Our main findings confirm the efficacy of our approach:
{\name}-PS outscores comparable methods under the official evaluation protocol and continues to benefit from additional inference compute scale.
Moreover, we observe that continual updates benefit memory augmentation over multiple attempts and that selecting a subset of memory for each problem is a crucial component to enable a memory to continually grow without overwhelming the LLM context.
This paper's work represents early attempts toward the main tenets of higher-level abstraction and modularity. 
Promising future directions include hierarchical designs and consolidation mechanisms that restructure memory.
To encourage further investigation, we also release a concept-annotation dataset and configurable puzzle synthesis pipeline, providing resources for evaluating concept representations and advancing abstraction-based memory methods.

\section*{Acknowledgements}
We thank Rahman Hajiyev and Jiayu Liu for early contributions to this project.


\bibliography{iclr2026_conference}

\begin{thebibliography}{35}
\providecommand{\natexlab}[1]{#1}
\providecommand{\url}[1]{\texttt{#1}}
\expandafter\ifx\csname urlstyle\endcsname\relax
  \providecommand{\doi}[1]{doi: #1}\else
  \providecommand{\doi}{doi: \begingroup \urlstyle{rm}\Url}\fi

\bibitem[Akyürek et~al.(2025)Akyürek, Damani, Zweiger, Qiu, Guo, Pari, Kim, and Andreas]{akyurek2024ttt}
Ekin Akyürek, Mehul Damani, Adam Zweiger, Linlu Qiu, Han Guo, Jyothish Pari, Yoon Kim, and Jacob Andreas.
\newblock The surprising effectiveness of test-time training for few-shot learning, 2025.
\newblock URL \url{https://arxiv.org/abs/2411.07279}.

\bibitem[ARC-Prize(2025)]{arcprizeleaderboard}
ARC-Prize, 2025.
\newblock URL \url{https://arcprize.org/leaderboard}.

\bibitem[Asai et~al.(2023)Asai, Wu, Wang, Sil, and Hajishirzi]{asai2023selfrag}
Akari Asai, Zeqiu Wu, Yizhong Wang, Avirup Sil, and Hannaneh Hajishirzi.
\newblock Self-rag: Learning to retrieve, generate, and critique through self-reflection, 2023.
\newblock URL \url{https://arxiv.org/abs/2310.11511}.

\bibitem[Baek et~al.(2024)Baek, Jauhar, Cucerzan, and Hwang]{Baek2024ResearchAgentIR}
Jinheon Baek, Sujay~Kumar Jauhar, Silviu Cucerzan, and Sung~Ju Hwang.
\newblock Researchagent: Iterative research idea generation over scientific literature with large language models.
\newblock \emph{ArXiv}, abs/2404.07738, 2024.
\newblock URL \url{https://api.semanticscholar.org/CorpusID:269042844}.

\bibitem[Bottou \& LeCun(2003)Bottou and LeCun]{bottou2003large}
L{\'e}on Bottou and Yann LeCun.
\newblock Large scale online learning.
\newblock \emph{Advances in neural information processing systems}, 16, 2003.

\bibitem[Bottou \& LeCun(2005)Bottou and LeCun]{bottou2005line}
L{\'e}on Bottou and Yann LeCun.
\newblock On-line learning for very large data sets.
\newblock \emph{Applied stochastic models in business and industry}, 21\penalty0 (2):\penalty0 137--151, 2005.

\bibitem[Bulatov et~al.(2022)Bulatov, Kuratov, and Burtsev]{bulatov2022recurrentmemorytransformer}
Aydar Bulatov, Yuri Kuratov, and Mikhail~S. Burtsev.
\newblock Recurrent memory transformer, 2022.
\newblock URL \url{https://arxiv.org/abs/2207.06881}.

\bibitem[Chhikara et~al.(2025)Chhikara, Khant, Aryan, Singh, and Yadav]{mem0}
Prateek Chhikara, Dev Khant, Saket Aryan, Taranjeet Singh, and Deshraj Yadav.
\newblock Mem0: Building production-ready ai agents with scalable long-term memory.
\newblock \emph{arXiv preprint arXiv:2504.19413}, 2025.

\bibitem[Chollet(2019)]{chollet2019measureintelligence}
François Chollet.
\newblock On the measure of intelligence, 2019.
\newblock URL \url{https://arxiv.org/abs/1911.01547}.

\bibitem[Fang et~al.(2025)Fang, Liang, Wang, Wu, Qiao, Xie, Huang, Chen, and Zhang]{fang2025memp}
Runnan Fang, Yuan Liang, Xiaobin Wang, Jialong Wu, Shuofei Qiao, Pengjun Xie, Fei Huang, Huajun Chen, and Ningyu Zhang.
\newblock Memp: Exploring agent procedural memory, 2025.
\newblock Work in progress.

\bibitem[Gao et~al.(2025)Gao, Geng, Hua, Hu, Juan, Liu, Liu, Qiu, Qi, Wu, Wang, Xiao, Zhou, Zhang, Zhang, Xiang, Fang, Zhao, Liu, Ren, Qian, Wang, Hu, Wang, Wu, Ji, and Wang]{gao2025selfevolvesurvey}
Huanang Gao, Jiayi Geng, Wenyue Hua, Mengkang Hu, Xinzhe Juan, Hongzhang Liu, Shilong Liu, Jiahao Qiu, Xuan Qi, Yiran Wu, Hongru Wang, Han Xiao, Yuhang Zhou, Shaokun Zhang, Jiayi Zhang, Jinyu Xiang, Yixiong Fang, Qiwen Zhao, Dongrui Liu, Qihan Ren, Cheng Qian, Zhenhailong Wang, Minda Hu, Huazheng Wang, Qingyun Wu, Heng Ji, and Mengdi Wang.
\newblock A survey of self-evolving agents: On path to artificial super intelligence, 2025.
\newblock URL \url{https://arxiv.org/abs/2507.21046}.

\bibitem[He et~al.(2024{\natexlab{a}})He, Li, Xing, Li, Tang, and Ding]{He2024MakeLB}
Pengfei He, Zitao Li, Yue Xing, Yaling Li, Jiliang Tang, and Bolin Ding.
\newblock Make llms better zero-shot reasoners: Structure-orientated autonomous reasoning.
\newblock \emph{ArXiv}, abs/2410.19000, 2024{\natexlab{a}}.
\newblock URL \url{https://api.semanticscholar.org/CorpusID:273638104}.

\bibitem[He et~al.(2024{\natexlab{b}})He, Karlinsky, Kim, McAuley, Krotov, and Feris]{he2024camelotlargelanguagemodels}
Zexue He, Leonid Karlinsky, Donghyun Kim, Julian McAuley, Dmitry Krotov, and Rogerio Feris.
\newblock Camelot: Towards large language models with training-free consolidated associative memory, 2024{\natexlab{b}}.
\newblock URL \url{https://arxiv.org/abs/2402.13449}.

\bibitem[Hu et~al.(2024)Hu, Chen, Chen, Mu, Shao, and Luo]{hu2024hiagent}
Mengkang Hu, Tianxing Chen, Qiguang Chen, Yao Mu, Wenqi Shao, and Ping Luo.
\newblock Hiagent: Hierarchical working memory management for solving long-horizon agent tasks with large language model.
\newblock \emph{arXiv preprint arXiv:2408.09559}, 2024.

\bibitem[Huang et~al.(2024)Huang, Chen, Mishra, Zheng, Yu, Song, and Zhou]{huang2024cannotcorrect}
Jie Huang, Xinyun Chen, Swaroop Mishra, Huaixiu~Steven Zheng, Adams~Wei Yu, Xinying Song, and Denny Zhou.
\newblock Large language models cannot self-correct reasoning yet, 2024.
\newblock URL \url{https://arxiv.org/abs/2310.01798}.

\bibitem[Kamoi et~al.(2024)Kamoi, Zhang, Zhang, Han, and Zhang]{kamoi2024correctmistakes}
Ryo Kamoi, Yusen Zhang, Nan Zhang, Jiawei Han, and Rui Zhang.
\newblock When can llms actually correct their own mistakes? a critical survey of self-correction of llms, 2024.
\newblock URL \url{https://arxiv.org/abs/2406.01297}.

\bibitem[Lewis et~al.(2021)Lewis, Perez, Piktus, Petroni, Karpukhin, Goyal, Küttler, Lewis, tau Yih, Rocktäschel, Riedel, and Kiela]{lewis2021retrievalaugmentedgenerationknowledgeintensivenlp}
Patrick Lewis, Ethan Perez, Aleksandra Piktus, Fabio Petroni, Vladimir Karpukhin, Naman Goyal, Heinrich Küttler, Mike Lewis, Wen tau Yih, Tim Rocktäschel, Sebastian Riedel, and Douwe Kiela.
\newblock Retrieval-augmented generation for knowledge-intensive nlp tasks, 2021.
\newblock URL \url{https://arxiv.org/abs/2005.11401}.

\bibitem[Li et~al.(2024)Li, Hu, Larsen, Wu, Alford, Woo, Dunn, Tang, Naim, Nguyen, Zheng, Tavares, Pu, and Ellis]{li2024barc}
Wen-Ding Li, Keya Hu, Carter Larsen, Yuqing Wu, Simon Alford, Caleb Woo, Spencer~M. Dunn, Hao Tang, Michelangelo Naim, Dat Nguyen, Wei-Long Zheng, Zenna Tavares, Yewen Pu, and Kevin Ellis.
\newblock Combining induction and transduction for abstract reasoning, 2024.
\newblock URL \url{https://arxiv.org/abs/2411.02272}.

\bibitem[Liu et~al.(2023)Liu, Yang, Shen, Hu, Zhang, Gu, and Zhang]{liu2023thinkinmemoryrecallingpostthinkingenable}
Lei Liu, Xiaoyan Yang, Yue Shen, Binbin Hu, Zhiqiang Zhang, Jinjie Gu, and Guannan Zhang.
\newblock Think-in-memory: Recalling and post-thinking enable llms with long-term memory, 2023.
\newblock URL \url{https://arxiv.org/abs/2311.08719}.

\bibitem[Liu et~al.(2025)Liu, Si, Narasimhan, and Yao]{liu2025contextual}
Yitao Liu, Chenglei Si, Karthik Narasimhan, and Shunyu Yao.
\newblock Contextual experience replay for self-improvement of language agents.
\newblock \emph{arXiv preprint arXiv:2506.06698}, 2025.

\bibitem[Madaan et~al.(2023)Madaan, Tandon, Gupta, Hallinan, Gao, Wiegreffe, Alon, Dziri, Prabhumoye, Yang, Gupta, Majumder, Hermann, Welleck, Yazdanbakhsh, and Clark]{madaan2023selfrefine}
Aman Madaan, Niket Tandon, Prakhar Gupta, Skyler Hallinan, Luyu Gao, Sarah Wiegreffe, Uri Alon, Nouha Dziri, Shrimai Prabhumoye, Yiming Yang, Shashank Gupta, Bodhisattwa~Prasad Majumder, Katherine Hermann, Sean Welleck, Amir Yazdanbakhsh, and Peter Clark.
\newblock Self-refine: Iterative refinement with self-feedback, 2023.
\newblock URL \url{https://arxiv.org/abs/2303.17651}.

\bibitem[Modarressi et~al.(2024)Modarressi, K{\"o}ksal, Imani, Fayyaz, and Sch{\"u}tze]{modarressi2024memllm}
Ali Modarressi, Abdullatif K{\"o}ksal, Ayyoob Imani, Mohsen Fayyaz, and Hinrich Sch{\"u}tze.
\newblock Memllm: Finetuning llms to use an explicit read-write memory.
\newblock \emph{arXiv preprint arXiv:2404.11672}, 2024.

\bibitem[Packer et~al.(2024)Packer, Wooders, Lin, Fang, Patil, Stoica, and Gonzalez]{packer2024memgptllmsoperatingsystems}
Charles Packer, Sarah Wooders, Kevin Lin, Vivian Fang, Shishir~G. Patil, Ion Stoica, and Joseph~E. Gonzalez.
\newblock Memgpt: Towards llms as operating systems, 2024.
\newblock URL \url{https://arxiv.org/abs/2310.08560}.

\bibitem[Park et~al.(2023)Park, O'Brien, Cai, Morris, Liang, and Bernstein]{park2023generativeagentsinteractivesimulacra}
Joon~Sung Park, Joseph~C. O'Brien, Carrie~J. Cai, Meredith~Ringel Morris, Percy Liang, and Michael~S. Bernstein.
\newblock Generative agents: Interactive simulacra of human behavior, 2023.
\newblock URL \url{https://arxiv.org/abs/2304.03442}.

\bibitem[Pourcel et~al.(2025)Pourcel, Colas, and Oudeyer]{pourcel2025soararc}
Julien Pourcel, Cédric Colas, and Pierre-Yves Oudeyer.
\newblock Self-improving language models for evolutionary program synthesis: A case study on arc-agi, 2025.
\newblock URL \url{https://arxiv.org/abs/2507.14172}.

\bibitem[Qiu et~al.(2024)Qiu, Jiang, Lu, Sclar, Pyatkin, Bhagavatula, Wang, Kim, Choi, Dziri, and Ren]{qiu2024phenomenalpuzzlingtestinginductive}
Linlu Qiu, Liwei Jiang, Ximing Lu, Melanie Sclar, Valentina Pyatkin, Chandra Bhagavatula, Bailin Wang, Yoon Kim, Yejin Choi, Nouha Dziri, and Xiang Ren.
\newblock Phenomenal yet puzzling: Testing inductive reasoning capabilities of language models with hypothesis refinement, 2024.
\newblock URL \url{https://arxiv.org/abs/2310.08559}.

\bibitem[Shinn et~al.(2023)Shinn, Cassano, Berman, Gopinath, Narasimhan, and Yao]{shinn2023reflexion}
Noah Shinn, Federico Cassano, Edward Berman, Ashwin Gopinath, Karthik Narasimhan, and Shunyu Yao.
\newblock Reflexion: Language agents with verbal reinforcement learning, 2023.
\newblock URL \url{https://arxiv.org/abs/2303.11366}.

\bibitem[Suzgun et~al.(2025)Suzgun, Yuksekgonul, Bianchi, Jurafsky, and Zou]{suzgun2025dynamiccheatsheettesttimelearning}
Mirac Suzgun, Mert Yuksekgonul, Federico Bianchi, Dan Jurafsky, and James Zou.
\newblock Dynamic cheatsheet: Test-time learning with adaptive memory, 2025.
\newblock URL \url{https://arxiv.org/abs/2504.07952}.

\bibitem[Wang et~al.(2025)Wang, Li, Sun, Chen, Liu, Wu, Lu, Song, and Yadkori]{wang2025hrm}
Guan Wang, Jin Li, Yuhao Sun, Xing Chen, Changling Liu, Yue Wu, Meng Lu, Sen Song, and Yasin~Abbasi Yadkori.
\newblock Hierarchical reasoning model, 2025.
\newblock URL \url{https://arxiv.org/abs/2506.21734}.

\bibitem[Wang et~al.(2023{\natexlab{a}})Wang, Xie, Jiang, Mandlekar, Xiao, Zhu, Fan, and Anandkumar]{wang2023voyageropenendedembodiedagent}
Guanzhi Wang, Yuqi Xie, Yunfan Jiang, Ajay Mandlekar, Chaowei Xiao, Yuke Zhu, Linxi Fan, and Anima Anandkumar.
\newblock Voyager: An open-ended embodied agent with large language models, 2023{\natexlab{a}}.
\newblock URL \url{https://arxiv.org/abs/2305.16291}.

\bibitem[Wang et~al.(2024)Wang, Zelikman, Poesia, Pu, Haber, and Goodman]{wang2024hypothesissearchinductivereasoning}
Ruocheng Wang, Eric Zelikman, Gabriel Poesia, Yewen Pu, Nick Haber, and Noah~D. Goodman.
\newblock Hypothesis search: Inductive reasoning with language models, 2024.
\newblock URL \url{https://arxiv.org/abs/2309.05660}.

\bibitem[Wang et~al.(2023{\natexlab{b}})Wang, Yang, Qiu, Liang, He, Gu, Xiao, and Wang]{Wang2023KnowledGPTEL}
Xintao Wang, Qian Yang, Yongting Qiu, Jiaqing Liang, Qi~He, Zhouhong Gu, Yanghua Xiao, and W.~Wang.
\newblock Knowledgpt: Enhancing large language models with retrieval and storage access on knowledge bases.
\newblock \emph{ArXiv}, abs/2308.11761, 2023{\natexlab{b}}.
\newblock URL \url{https://api.semanticscholar.org/CorpusID:261076315}.

\bibitem[Wu et~al.(2022)Wu, Rabe, Hutchins, and Szegedy]{wu2022memorizingtransformers}
Yuhuai Wu, Markus~N. Rabe, DeLesley Hutchins, and Christian Szegedy.
\newblock Memorizing transformers, 2022.
\newblock URL \url{https://arxiv.org/abs/2203.08913}.

\bibitem[Yang et~al.(2024)Yang, Yu, Zhang, Cao, Xu, Zhang, Gonzalez, and Cui]{yang2024bufferthoughtsthoughtaugmentedreasoning}
Ling Yang, Zhaochen Yu, Tianjun Zhang, Shiyi Cao, Minkai Xu, Wentao Zhang, Joseph~E. Gonzalez, and Bin Cui.
\newblock Buffer of thoughts: Thought-augmented reasoning with large language models, 2024.
\newblock URL \url{https://arxiv.org/abs/2406.04271}.

\bibitem[Zhang et~al.(2019)Zhang, Han, Liu, Jiang, Sun, and Liu]{zhang2019ernie}
Zhengyan Zhang, Xu~Han, Zhiyuan Liu, Xin Jiang, Maosong Sun, and Qun Liu.
\newblock Ernie: Enhanced language representation with informative entities, 2019.
\newblock URL \url{https://arxiv.org/abs/1905.07129}.

\end{thebibliography}
\bibliographystyle{iclr2026_conference}

\appendix

\section{Implementation Details}
\label{appendix:impl_details}
\subsection{PS Memory Format}
\label{appendix:ps_mem_fields}
Each PS format memory entry contains:
\begin{itemize}
    \item \textbf{Title}: a succinct label for the underlying idea.
    \item \textbf{Description}: elaboration on behavior and role.
    \item \textbf{Kind}: whether this concept encodes a type/structure/routine.
    \item \textbf{Parameters}: a list of fields that parametrize this concept.
    \item \textbf{Output Typing}: specifies what the output for this routine is, to suggest how various routines can plug into other routines.
    \item \textbf{Relevance Cues}: much like the \texttt{situation} field in the OE concepts, we consider the context in which this concept is relevant.
    \item \textbf{Implementation Notes}: suggestions on how to implement the concept in actual code.
\end{itemize}

\subsection{Experiment Parameters}
\label{appendix:params}
To build on frontier models, we experiment primarily with OpenAI's o4-mini (max tokens=32000, reasoning\_effort=medium).
For auxiliary tasks such as concept abstraction and non-reasoning selection, we use OpenAI's GPT-4.1 (temperature=0.3, max tokens=1000) to reduce token usage.
We looked into evaluating the open-source DeepSeek R1, which also has a transparent thinking process, but the output 8000 token limit consistently yielded unfinished solutions.

\section{Evaluation Details}
\label{appendix:eval_details}
We use ARC-AGI's official scoring metric, which we term oracle@$k$ (each test case is scored separately, and if any of $k$ candidates pass, full credit is given for that test case).
The ARC-AGI official evaluation has $k=2$, but to mitigate sampling variance, we sample 3 runs for each setting and report the average single run score, average oracle@2 score, and their standard deviation.
We investigate several alternate scoring protocols (requiring a single program to solve all test cases, accumulating test case solve rates, and requiring references also to be solved) and include the results in \autoref{tab:main_strict}.

Here is a precise definition of our scoring procedure for a single problem $P$. 
Let...
\begin{itemize}
\item $X$ be the size $n$ set of attempts on problem $P$
\item $T$ be the set of test cases for problem $P$
\item $C$ represent a particular $k$-subset of $X$, in notation: $C \in [X]^{k}$
\end{itemize}
\[score = \frac{1}{|[X]^{k}|}  \sum\limits_{C \in [X]^{k}} z(C)\]
where $z$ is the problem score by a $k$-attempt ensemble
\[z(C) = \frac{1}{|T|} \sum\limits_{t_{i} \in T} \mathds{1}\{\exists c \in C \:\: \texttt{Verify}(c, t_{i}) \}\]
where 
\begin{itemize}
\item $c$ is a single program attempt in the k-attempt ensemble, 
\item t is a test case,
\item and $\texttt{Verify}$ returns true if the program successfully passes the test case.
\end{itemize}

\begin{table}[ht]
\centering
\resizebox{\textwidth}{!}{%
\begin{tabular}{llrrr|lcr}
\toprule
\multicolumn{2}{c}{} & \multicolumn{3}{c}{Individual Run Scores} & \multicolumn{3}{c}{Oracle@k Scores} \\
\midrule
 &  & run0 & run1 & run2 & k=1 & \textbf{k=2} (Official Score) & k=3 \\
iteration & setting &  &  &  &  &  &  \\
\midrule
\multirow[t]{4}{*}{0} & baseline & 46.00 & 45.50 & 47.50 & 46.33 (1.04) & 55.17 (3.18) & 59.50 \\
 & cheatsheet & 48.00 & 44.50 & 50.00 & 47.50 (2.78) & 57.67 (2.52) & \textbf{64.00} \\
 & \name-OE (ours) & 49.00 & 47.00 & 48.00 & 48.00 (1.00) & 56.67 (1.53) & 61.00 \\
 & \name-PS (ours) & 49.50 & 49.00 & 49.50 & \textbf{49.33 (0.29)} & \textbf{59.33 (0.29)} & 63.50 \\
\cline{1-8}
\multirow[t]{4}{*}{1} & baseline & 57.00 & 55.50 & 61.00 & 57.83 (2.84) & 66.67 (3.82) & 71.50 \\
 & cheatsheet & 56.00 & 56.50 & 57.50 & 56.67 (0.76) & 65.67 (1.44) & 70.50 \\
 & \name-OE (ours) & 59.00 & 55.00 & 56.00 & 56.67 (2.08) & 65.67 (1.53) & 70.00 \\
 & \name-PS (ours) & 58.50 & 60.00 & 55.50 & \textbf{58.00 (2.29)} & \textbf{67.33 (1.61)} & \textbf{72.50} \\
\cline{1-8}
\multirow[t]{4}{*}{2} & baseline & 59.50 & 59.00 & 65.00 & 61.17 (3.33) & 69.00 (2.65) & 73.00 \\
 & cheatsheet & 65.00 & 61.00 & 61.00 & \textbf{62.33 (2.31)} & \textbf{71.33 (1.53)} & \textbf{76.00} \\
 & \name-OE (ours) & 62.00 & 59.00 & 61.00 & 60.67 (1.53) & 67.67 (2.52) & 71.00 \\
 & \name-PS (ours) & 60.50 & 66.00 & 58.50 & 61.67 (3.88) & 70.83 (3.06) & 75.50 \\
\cline{1-8}
\bottomrule
\end{tabular}
}
\caption{\textbf{Full o4-mini Results.} The right partition of the table containing aggregate scores shows the inference scaling: going down represents sequential retry, going right represents adding parallel samples. {\name}-PS is the only setting that outperforms the baseline consistently at every tested scale.}
\label{tab:full_res}
\end{table}

\begin{table}[ht]
\centering
\resizebox{\textwidth}{!}{%
\small
\begin{tabular}{llrrrllr}
\toprule
 &  & run0 & run1 & run2 & k=1 & k=2 & k=3 \\
iteration & setting &  &  &  &  &  &  \\
\midrule
\multirow[t]{4}{*}{0} & baseline & 46.00 & 45.00 & 47.00 & 46.00 (1.00) & 54.67 (3.21) & 59.00 \\
 & cheatsheet & 48.00 & 44.00 & 50.00 & 47.33 (3.06) & 57.67 (2.52) & 64.00 \\
 & \name-OE & 49.00 & 47.00 & 48.00 & 48.00 (1.00) & 56.67 (1.53) & 61.00 \\
 & \name-PS & 49.00 & 49.00 & 49.00 & 49.00 (0.00) & 59.00 (0.00) & 63.00 \\
\cline{1-8}
\multirow[t]{4}{*}{1} & baseline & 57.00 & 55.00 & 61.00 & 57.67 (3.06) & 66.33 (4.04) & 71.00 \\
 & cheatsheet & 56.00 & 56.00 & 57.00 & 56.33 (0.58) & 65.33 (1.15) & 70.00 \\
 & \name-OE & 59.00 & 55.00 & 56.00 & 56.67 (2.08) & 65.67 (1.53) & 70.00 \\
 & \name-PS & 58.00 & 60.00 & 55.00 & 57.67 (2.52) & 67.00 (1.73) & 72.00 \\
\cline{1-8}
\multirow[t]{4}{*}{2} & baseline & 59.00 & 59.00 & 65.00 & 61.00 (3.46) & 69.00 (2.65) & 73.00 \\
 & cheatsheet & 65.00 & 61.00 & 61.00 & 62.33 (2.31) & 71.33 (1.53) & 76.00 \\
 & \name-OE & 62.00 & 59.00 & 61.00 & 60.67 (1.53) & 67.67 (2.52) & 71.00 \\
 & \name-PS & 60.00 & 66.00 & 58.00 & 61.33 (4.16) & 70.33 (3.06) & 75.00 \\
\cline{1-8}
\bottomrule
\end{tabular}

}
\caption{
\textbf{Strict Scoring}.
While the official evaluation scheme allows different test cases solved by different attempts to be ensembled,
in contrast, the strict scoring regime only marks a puzzle as solved if a single attempt (generated program) solves all test cases.
}
\label{tab:main_strict}
\end{table}

\section{Further Discussion}
\label{appendix:further_discussion}

\begin{figure}
    \centering
    \includegraphics[width=0.8\linewidth]{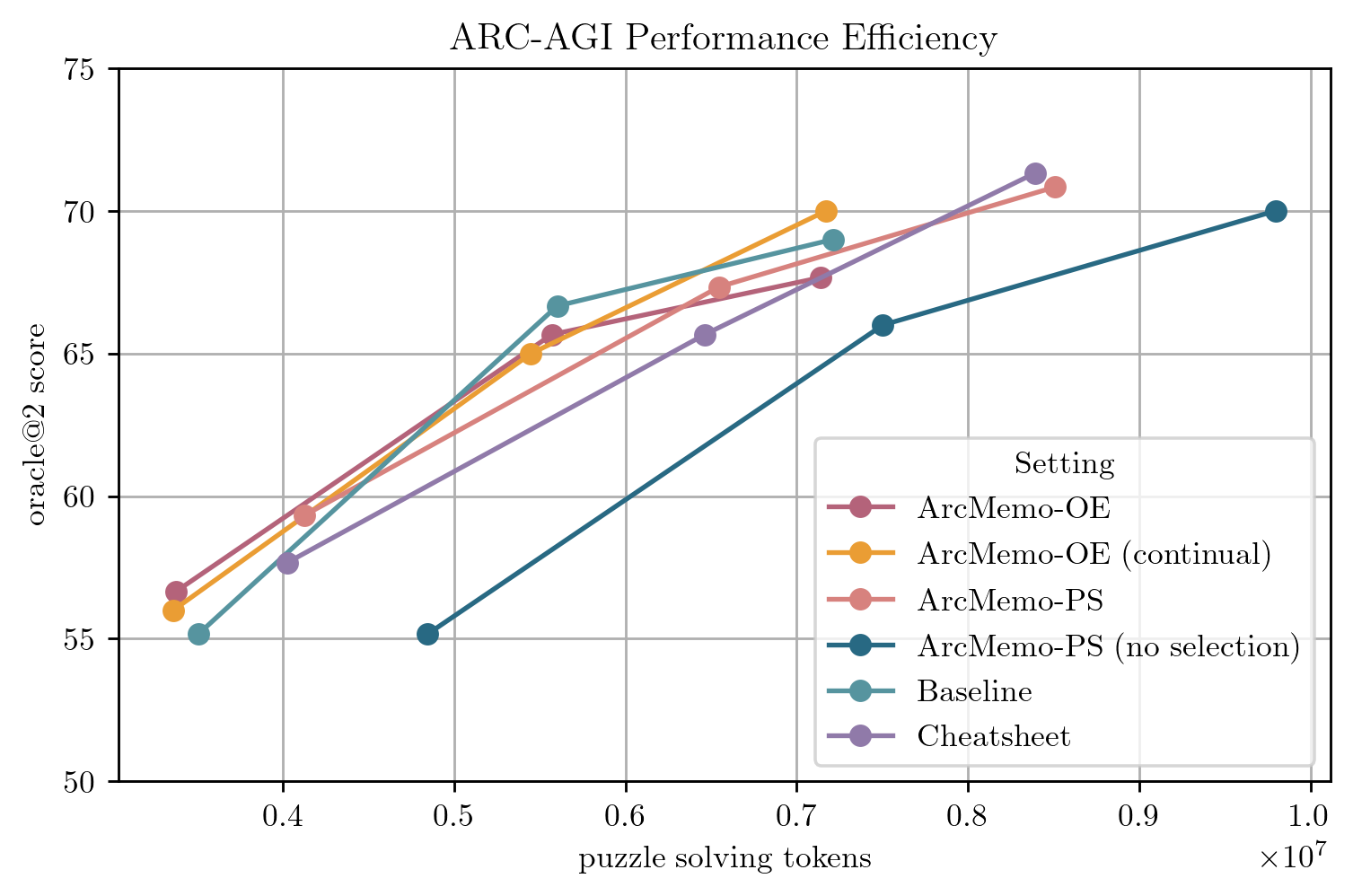}
    \caption{Token Efficiency Plot: Comparing various settings' official score (oracle@2) on the public validation subset to the tokens used by the reasoning model for strictly puzzle solving.}
    \label{fig:token_efficiency}
\end{figure}

\paragraph{Token Efficiency}
We find that our system's token efficiency is similar to the baseline.
We observe that memory-augmented runs tend to increase output token usage.
While this seems to counter the intuition that retrieving rather than regenerating certain ideas would save tokens, we hypothesize that introducing memory leads the model to explore more hypotheses.
In other words, selection inaccuracies (false positives) seem to be manageable by the reasoning models, but at the cost of increased token usage.
Full results are plotted in \autoref{fig:token_efficiency}. 

\paragraph{Concept Specificity Investigation}
As part of the development of our concept representations, we conducted small-scale experiments using tiny (n=10) subsets of the validation split to investigate the level of detail needed for concepts to be useful.
Manually writing maximum detail situation-suggestion style concepts and using iteratively LLM-summarized versions (for a total of 5 levels of specificity for each concept), we found the reasonably expected result that higher levels of specificity correspond to better solve rates. 
Maximum specificity solved 4/10 with a generally decreasing pattern with lower specificity. 
While this result was entirely expected, the main goal of the experiment was to determine how much we could compromise on concept detail to improve the retrieval aspect.
There is a tension between retrieval and puzzle solving in the sense that puzzle solving benefits from more detailed suggestions, but retrieval struggles to apply them to new situations when the concept's relevant situation is so precisely defined.

\paragraph{Embedding-based Retrieval Experiments}
Early in testing, we investigated standard embedding-based retrieval approaches.
Following the \name-OE setting, we used a VLM-generated caption to query a vector database of \name-OE-style concept embeddings.
While computationally cheap compared to autoregressive generation (especially so when leveraging long-form reasoning), we saw generally poor retrieval results.
Using OpenAI's embeddings API and an o3-mini puzzle-solving backbone proved ineffective, lowering the score from 0.26 to 0.22, marking a 15\% reduction.
Qualitative analysis found that retrieved concepts were largely irrelevant to the puzzle and seemed to over-index on lexical similarity.
Moreover, a small set of broadly defined (non-specific) concepts was found to be grossly overrepresented.

\section{Limitations and Future Work}
\label{appendix:limitations}


\subsection{Limitations}
Following \cite{akyurek2024ttt}, our evaluation focuses on a 100-task subset of ARC-AGI-1, selected to make large-scale experimentation feasible under compute and cost constraints. 
This choice was further motivated by the substantial sampling variance we observed—identical prompts can yield noticeably different scores, requiring multiple runs to obtain stable estimates. 
As a result, our study concentrates on a relatively small frontier of puzzles where memory augmentation can yield new solves, limiting the absolute magnitude of observable gains. 

In addition, ARC-AGI-2 was released mid-project as a harder successor benchmark, with state-of-the-art performance still below 30\%.  
Re-running all baselines and experiments on ARC-AGI-2 was infeasible within scope due to cost and time constraints, but extending memory-based approaches to ARC-AGI-2 remains an important direction for future work.  

\subsection{Future Work}
Several directions follow naturally from this work. One is to broaden evaluation to larger and more challenging puzzle sets, including those unsolved by the baseline, to test both new solve potential and consistency on borderline cases. Another is to explore order-robust update strategies and curriculum designs that reduce sensitivity to problem ordering while preserving inter-problem learning benefits.  
Beyond these extensions, a key avenue is hierarchical abstraction: moving from additive concept updates toward consolidation operations that restructure the memory across multiple experiences. Finally, our current design choices for representation and retrieval were made under cost constraints; richer optimization may yield substantial improvements.  

To facilitate further research, \textbf{we release a data resource}: hand-annotated concepts for difficult puzzles, together with a configurable puzzle synthesis pipeline. This resource is intended to support isolated evaluation of concept representations and selection mechanisms, as well as development of methods for concept abstraction. We hope it provides a useful foundation for continued exploration of abstract reasoning memory.

\section{Additional Related Work}
\label{appendix:related_work}
\paragraph{Knowledge Base Augmented LLMs for Factuality.}
Early efforts to enhance language models with external memory primarily targeted knowledge-intensive tasks. Retrieval-Augmented Generation (RAG) introduced the idea of retrieving documents from an external store based on embedding similarity to augment generation \citep{lewis2021retrievalaugmentedgenerationknowledgeintensivenlp}.
Follow-up work like KnowledGPT incorporated formal knowledge bases, allowing LLMs to execute structured queries through ``Program-of-Thought'' prompting for multi-hop reasoning \citep{Wang2023KnowledGPTEL}. 
More recently, MemLLM proposed inline memory operations—allowing the model to read from and write to memory directly during generation—enabling continual adaptation without re-training \citep{modarressi2024memllm}.

\paragraph{Embedding Space Memory.} 
Another stream of work aims to overcome the limited context window of transformers through architectural interventions that compress information from long sequences into continuous space.
Memorizing Transformers stored latent key-value vectors from past inputs and retrieved them via a kNN mechanism to be attended to alongside the current context \citep{wu2022memorizingtransformers}. 
The Recurrent Memory Transformer extended this by incorporating a writable memory updated over time, making memory content responsive to ongoing computation \citep{bulatov2022recurrentmemorytransformer}. 
\cite{he2024camelotlargelanguagemodels} blended these ideas by applying a moving average to update memory slots, allowing for both similarity-based retrieval and continual adaptation \cite{He2024MakeLB}. 
MemGPT introduced a hierarchical memory system with a working (RAM-like) memory and a larger archival memory, simulating long-term information management in a more structured fashion \citep{packer2024memgptllmsoperatingsystems}.

\paragraph{Parameter-Free Test-time Learning.}
Test-time learning (sometimes referred to as online learning) describes methods that update predictions at inference time \cite{bottou2003large,bottou2005line}.
However, trends towards (1) massive model sizes and (2) black box models accessed through APIs make continually updating parameters computationally impractical or downright impossible.
In light of this, LLM test-time learning research sought to pursue parameter-free adaptation by using LLMs' remarkable capacity for instruction following and self-reflection.
Methods such as Reflexion \citep{shinn2023reflexion}, Self-Refine \citep{madaan2023selfrefine}, and Self-RAG \citep{asai2023selfrag}, \textit{inter alia} attempt to correct mistakes at test-time by way of self-reflection.
Follow-up work \citep{huang2024cannotcorrect,kamoi2024correctmistakes} finds that self-correction methods depend on the availability of some extrinsic feedback or verification mechanism to reliably improve end performance and that these methods produce dubious results absent these grounded sources of feedback.
Still, self-reflection remains a key component of self-evolving agents \citep{gao2025selfevolvesurvey}.

\paragraph{Memory for Agentic Skill Acquisition.}
Memory also plays a crucial role in LLM-based agents. 
\cite{park2023generativeagentsinteractivesimulacra} structured memory as a stream of episodic observations and distilled high-level summaries from them to guide decision-making.
Contextual Experience Replay discretizes trajectories into ``experiences'' (containing environment dynamics from a trajectory and skills storing actions from said trajectory) and retrieves relevant blocks to guide new episodes \citep{liu2025contextual}. 
HiAgent organizes working memory (within a single query) hierarchically by subgoals, summarizing and replacing traces to improve long-horizon efficiency \citep{hu2024hiagent}. 
External-memory systems such as Mem0 and MemP manage a persistent store with explicit add–update–prune operations (MemP casts control as an MDP), yielding gains on dialogue and tool-use/planning tasks \citep{mem0,fang2025memp}.
MemP in particular is a concurrent work that also demonstrates the efficacy of abstraction, although it targets rewriting action sequences as step sequences in explicitly agentic settings, contrasting with {\name}'s general problem-solving focus. 
Voyager uses LLMs to autonomously acquire and store skills while exploring Minecraft, with a growing library of reusable code snippets written by the agent itself \cite{wang2023voyageropenendedembodiedagent}. 
Other systems, like the ResearchAgent, use memory initialized from domain-specific corpora, such as scientific papers, to ground query responses in high-utility prior knowledge \citep{Baek2024ResearchAgentIR}.

\end{document}